\documentclass[runningheads]{llncs}
\usepackage[T1]{fontenc}
\usepackage{makecell}
\usepackage{amsmath}

\usepackage{tabularx}

\usepackage{pgfplots}

\pgfplotsset{compat=1.18}

\usepackage[caption=false]{subfig}

\usepackage{pgfplots}

\pgfplotsset{compat=1.18}
\usetikzlibrary{positioning}
\usetikzlibrary{arrows.meta, positioning}

\begin{document}
\title{Evaluation of Population Initialization Methods for Genetic Programming-based Symbolic Regression}
\titlerunning{Evaluation of Population Initialization Methods}
\author{
  Lukas Kammerer\inst{1}\orcidID{0000-0001-8236-4294} 
  \and Gabriel Kronberger\inst{1}\orcidID{0000-0002-3012-3189}
  \and Deaglan J. Bartlett\inst{2}\orcidID{0000-0001-9426-7723} 
  \and Harry Desmond\inst{3}\orcidID{0000-0003-0685-9791} 
  \and Pedro G. Ferreira\inst{2}\orcidID{0000-0002-3021-2851}
  \and Stephan Winkler\inst{1}\orcidID{0000-0002-5196-4294}
}
\authorrunning{L.~Kammerer et al.}
\institute{
  Heuristic and Evolutionary Algorithms Laboratory, \\
  University of Applied Sciences Upper Austria, Hagenberg, Austria \\
  \and
  Astrophysics, University of Oxford, Oxford, UK \\
  \and
  Institute of Cosmology \& Gravitation, University of Portsmouth, Portsmouth, UK
  \medskip
  \email{lukas.kammerer@fh-hagenberg.at}
}
\maketitle
\begin{abstract}
We analyze the effect of optimizing the initial population of genetic programming (GP) for symbolic regression (SR) on the accuracy and complexity of solutions. We compare three well-established random initialization methods as well as initialization with small optimized solutions from exhaustive symbolic regression (ESR) using a GP/SR implementation which is based on the multi-objective evolutionary algorithm NSGA-II. We compare the final Pareto fronts found with each initialization method on twelve synthetic problems of varying complexity and one real-world dataset. 
We find no significant differences in accuracy or model complexity among the initialization methods. The initial advantage of initialization with ESR  disappears after only a few generations.
Our results show that, given similar diversity in the initial population, the effect of the initialization method in GP-based symbolic regression on the final Pareto front is negligible.

\end{abstract}
\section{Introduction}

Symbolic regression (SR) is a machine learning problem where the goal is to find a human-interpretable mathematical expression of any possible functional form that best fits a given dataset \cite{koza1992,kronberger2024symbolic}. The most common approach to solve SR problems is genetic programming (GP), which evolves a population of individuals in the form of expression trees representing mathematical expressions. GP typically initializes a population randomly 
and then evolves it using selection, crossover, and mutation operations. This random initialization ensures a diverse set of individuals that can be used by the crossover operator to create new and potentially better models \cite{koza1992}. 

Most research on improving GP/SR towards more accurate or shorter models has focused on the evolutionary process itself, for example by introducing new selection, crossover, and mutation operators \cite{lacava2021contemporary}. Methods for creating an initial population and therefore initializing the starting condition of the evolutionary process are referred to as population seeding. These aim to provide aspects in the initial population that are beneficial for the evolutionary process, such as diversity \cite{paul2014new,hassanat2018improved}. Given the crucial role of population diversity for the success of GP, most developments on population initialization focus on creating a uniform distribution of symbols or number of nodes in expression trees, such as the so-called \textit{grow} method, the \textit{full} method, the combination of the latter called the ramped half-and-half method \cite{koza1992}, the probabilistic tree creators PTC-1 and PTC-2 \cite{luke2002two} or the balanced tree creator (BTC) \cite{burlacu2020operon}.

Different approaches to population seeding exist to deliberately bias initial populations. Methods that optimize the initial population's fitness showed clear improvements in combinatorial optimization \cite{paul2014new,deng2015improved,hassanat2018improved}, while related attempts in GP for program synthesis \cite{ahmad2018comparison} did not provide significant improvements over random initialization. Another approach by Langdon and Nordin \cite{langdon2000seeding} took overfit individuals as the initial population and used GP with parsimony pressure to identify well-generalizing models. Mundhenk et al.~\cite{mundhenk2021symbolic} used a recurrent neural network (RNN) to generate an initial population. After short GP runs that were seeded with the RNN, the best final models were then used to further train the RNN and generate a new population.

This work analyses the effects of optimizing the initial population versus well-established random initialization methods. 
Our main hypothesis is that an initial population that is already strong in accuracy and complexity either accelerates convergence or improves the final Pareto front, provided that it retains sufficient diversity. 

To check this, we use short, well-fitting expressions produced by exhaustive symbolic regression (ESR) \cite{bartlett2023exhaustive} for the initial GP population instead of random seeding. ESR performs a brute-force search over the space of all possible algebraically unique functional forms up to a certain number of nodes in the expression tree, which is referred to as complexity. Parameters are optimized numerically, and we assume in our experiments that the best parameters are found for each functional form. Due to its exhaustive nature, ESR is limited to a much smaller search space than GP/SR and covers only short, univariate models of complexity up to ten. We use the GP/SR implementation Operon\footnote{specifically the Python wrapper \textit{PyOperon}, version 0.6.0.} \cite{burlacu2020operon} which has shown strong performance on SR benchmarks \cite{lacava2021contemporary}. 

As baselines for comparison, we use the grow initialization method, PTC-2, and BTC. The grow method \cite{koza1992} creates trees of any complexity and shape within a given depth limit, which is the only available parameter to control the shape of the resulting trees. Due to this lack of fine-grained model complexity control, the grow method might generate only small or imbalanced trees that lead to low diversity, or too large trees \cite{luke2002two,burke2003ramped}. PTC-2 \cite{luke2002two} creates trees from a customizable distribution of model complexity and operator occurrence. By default, Operon uses the balanced tree creator, BTC, which tends to build balanced trees of minimal depth and uniformly distributed complexities \cite{burlacu2020operon}.

The use of ESR for population seeding is motivated by its guarantee to find the most accurate and algebraically unique models in their shortest form within the given complexity limits \cite{bartlett2023exhaustive}. The population returned by ESR can be considered as an optimal population with respect to accuracy and complexity within Operon's initialization complexity bounds because Operon also uses a complexity limit of ten for random initialization. ESR identifies and removes algebraically duplicate models in its search and returns a set of mathematically unique models, 
which leads to syntactically diverse models, as we demonstrate below.

We compare ESR initialization with three different random initialization methods on twelve synthetic problems of varying complexity, and one real-world problem. Our results show that the initial advantage of the ESR-initialized population quickly diminishes during the evolutionary process within only a few generations. Therefore, there is no significant improvement in terms of accuracy and complexity. Only for simple problems, in which the ground truth is very close to the best model found by ESR, do we find a significant improvement with an ESR-initialized population over the random initial population. The experiments also show that all three random initialization methods perform equally well, which implies that the specific choice of random initialization method barely has an effect on the final Pareto front of models in GP/SR.

Section \ref{sec:methodology} describes the experimental setup, including the initialization methods, the GP settings and the benchmark problems. Section \ref{sec:results} first outlines preliminary results regarding diversity and accuracy in the initial population to ensure a proper experimental setup. It then shows the modeling results of the synthetic problems and verifies their implications on a real-world dataset. Section \ref{sec:conclusion} discusses and concludes our work.

\section{Methodology}
\label{sec:methodology}

ESR is run before Operon to use the resulting set of best models as the initial population for Operon. For each dataset we run the ESR fitting phase with the same precomputed set of expressions from \cite{bartlett2023functionsets} that uses arithmetic operators and the power function.
To match Operon's distribution of model complexity in the initial population, we run ESR with different maximum complexities. As we use a population size of 1000 in GP and only binary operators that result in odd complexities, we run ESR with maximum complexities of three, five, seven, and nine and take the 250 most accurate models from each run. We use duplicate models for complexity three and five as for these complexity values fewer than 250 unique models exist.

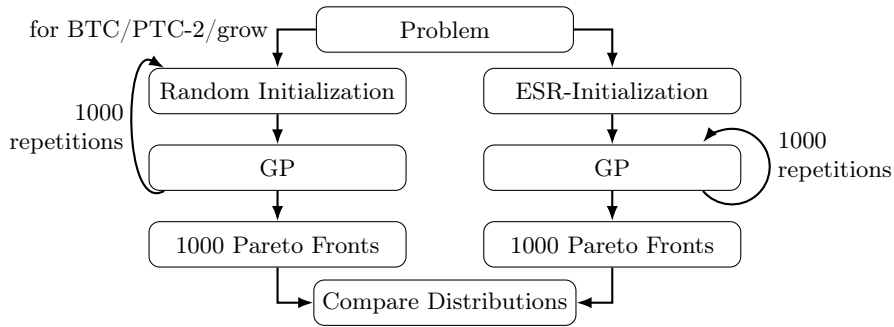
\begin{figure}[b!]
\begin{tikzpicture}[
    node distance=1cm,
    box/.style={draw, rectangle, rounded corners, minimum width=3.4cm, minimum height=0.6cm, align=center, font=\footnotesize},
    decision/.style={draw, diamond, aspect=2, minimum width=3cm, minimum height=0.7cm, align=center, font=\footnotesize},
    arrow/.style={-{Latex[length=2mm]}, thick}
]
    
\node[box] (start) {Problem};

\node[box, below left=0.2cm and -1.2cm of start] (method1) {Random Initialization};
\node[box, below right=0.2cm and -1.2cm of start] (method2) {ESR-Initialization};

\draw[arrow] (start.west) -| (method1.north) node[midway, above, font=\footnotesize, xshift=-1.7cm, yshift=-0.3cm] {for BTC/PTC-2/grow};
\draw[arrow] (start.east) -| (method2.north);

\node[box, below=0.4cm of method1] (runs1) {GP};
\node[box, below=0.4cm of method2] (runs2) {GP};

\draw[arrow] (method1) -- (runs1);
\draw[arrow] (method2) -- (runs2);

\draw [arrow] ([xshift=-1.5cm] runs1.south) to [out=270,in=90, bend left=120] node[midway, left, align=right] {1000\\repetitions} ([xshift=-1.5cm] method1.north);

\draw [arrow] ([xshift=1.2cm] runs2.south) arc(-140:140:.5) node[pos=0.55, right, align=left] {1000\\repetitions};

\node[box, below=0.4cm of runs1] (pareto1) {1000 Pareto Fronts};
\node[box, below=0.4cm of runs2] (pareto2) {1000 Pareto Fronts};

\draw[arrow] (runs1) -- (pareto1);
\draw[arrow] (runs2) -- (pareto2);

\node[box, below=0.4cm of start, yshift=-2.6cm] (compare) {Compare Distributions};

\draw[arrow] (pareto1.south) |- (compare.west);
\draw[arrow] (pareto2.south) |- (compare.east);
    
\end{tikzpicture}
\caption{We compare all initialization methods by running 1000 repetitions of GP with different random seeds on twelve synthetic and one real-world problem. We then compare the resulting Pareto fronts for each problem.}
\label{fig:setup}
\end{figure}

Given the deterministic nature of ESR (assuming perfect parameter optimization), we run it only once per problem. 
Since GP is a stochastic method, we perform 1000 repetitions of Operon with different random seeds for both ESR initialization and each random initialization method, as outlined in Figure \ref{fig:setup}. We note that the ESR optimization process is performed in addition to GP and therefore uses additional computational budget compared to GP with computationally cheap random initialization. We compare the models from all resulting 1000 Pareto fronts of each method and plot the distribution of the error of the Pareto-optimal models from all runs over their respective complexity for all initialization methods. By comparing all distributions, we show whether changing the initial population leads to improvements regarding both accuracy and complexity in the final Pareto fronts.

Since we are interested in the Pareto front of models over accuracy and complexity, we use the multi-objective NSGA-II \cite{deb2002fast} implementation of Operon. As objectives, we use the normalized mean squared error (NMSE) and the complexity, which is also referred to as the model length. The NMSE is defined as the mean squared error divided by the variance of the target variable in the training data. We use the same hyperparameters for all experiments, which are shown in Table~\ref{tab:operon_settings}, while only adapting the maximum complexity to specific problems. Problems 1 to 5 use a maximum complexity of 20, while Problems 6 to 12 use a maximum complexity of 50. The complexity limit for models in the initial population is set to ten, except for the grow method, which only supports a depth limit. Parameter values in randomly initialized models are sampled from a Gaussian distribution with zero mean and $\sigma=1$. In ESR-initialized populations, we retain the parameter values that were found by ESR in each model. During the evolutionary process, parameter values in each model are optimized with the Levenberg-Marquardt algorithm \cite{levenberg1944method,marquardt1963algorithm} for a maximum of 15 iterations. 

\begin{table}
  \centering
  \caption{GP/SR Hyperparameters used for all experiments. The maximum length depends on the problem. Other hyperparameters are set to their default values.
  }
  \label{tab:operon_settings}
  \begin{tabular}{l|l}
    \hline
    Hyperparameter & Value \\
    \hline
    Algorithm:          & NSGA-II \\
    Objectives:         & NMSE, Length \\
    Population Size:    & 1000 \\
    Generations:        & 500 \\
    Selection:          & Tournament (size=3) \\
    Mutation Rate:      & 25\% \\
    Max.~Length:        & 20 or 50 \\
    Max.~Initialization Length: & 10 (only ESR, PTC-2 and BTC) \\
    Max.~Initialization Depth:      & 5 \\
    Linear Scaling:     & off \\
    Variable Scaling:   & off \\
    Parameter Opt.:     & LM, max.~15 iterations \\
    Function Set:       & $+$, $-$, $\times$, $\div$, powerabs \\
    Terminal Set:       & $x$, parameters \\
  \hline
  \end{tabular}
\end{table}

The function set is based on the functions used by ESR, with the only difference being that univariate inverse operations are converted into binary division operations with one as the numerator. The function set consists of arithmetic operators and the composite \textit{powerabs} operator, defined as $\mathrm{powerabs}(a, b) = |a|^b$, which extends the power function to negative bases and ensures real-valued outputs for non-integer exponents. Operon's default linear and variable scaling was disabled to identify functions in their shortest representation and to reuse ESR's models in their original form. The most relevant parameters are shown in Table~\ref{tab:operon_settings}. All other parameters are set to their default values.

We use twelve synthetic problems with varying complexity and a small noise level for comparison, as shown in Table~\ref{tab:ground_truth_equations}. Each problem has a training set of size 100 and a test set of size 1000. 
We choose univariate synthetic benchmarks with known noise distributions since ESR currently supports only functions of one variable and so that we can evaluate the likelihood.
Potential univariate problems in common benchmarks \cite{lacava2021contemporary} are too simple or lack noise specifications. Problems 1 to 6 were created by hand by combining randomly sampled ESR solutions. Problems 7 to 12 were created by letting Operon with the same hyperparameters as in Table \ref{tab:operon_settings} with a maximum length of 50 over-fit 
to ten points of Gaussian random noise with zero mean and $\sigma = 1$
and then selecting and simplifying the best-fitting continuous function as ground truth. The values of $x$ are evenly spaced in the given range to prevent bias from specific training or test set samples. The value range of $x$ was chosen so that all values $f(x)$ are on a similar scale. Small Gaussian noise with zero mean and $\sigma = 10^{-7}$ 
is added to the target variable $f(x)$ to focus on the optimization process rather than model selection.

\renewcommand{\arraystretch}{1.5}
\begin{table}[t!]
\centering
\footnotesize
\caption{Ground truth equations, their complexity 
and value ranges for the twelve synthetic problems. Gaussian noise $\mathcal{N}(0, 10^{-7})$ is added to $f(x)$.}
\label{tab:ground_truth_equations}
\begin{tabular}{c|c|l|l}
Nr. & Complexity & Ground truth $f(x)$ & $x \in$ \\
\hline
1  & 11 & $\frac{1}{x + x^{{0.05}^{x} - x}}$ & $[0.6, 2.5]$ \\
2  & 13 & $- 0.4 x \left(x^{2} - 0.5\right) + x^{x}$ & $[0.1, 1.7]$ \\
3  & 13 & $\frac{{0.16}^{- \frac{1}{x}} x^{x}}{x + 3}$ & $[0.6, 3]$ \\
4  & 11 & $x^{x - x^{3.5}} + \frac{1}{x}$ & $[0.5, 2.5]$ \\
5  & 13 & $0.1 x + \left(\frac{0.17}{{0.1}^{x} + 0.25}\right)^{x}$ & $[0.5, 2.5]$ \\
6  & 17 & $0.1 x + \left(\frac{0.17}{{0.1}^{x} + 0.3}\right)^{x} - \frac{0.05}{x}$ & $[0.2, 4]$ \\
7  & 41 & $\frac{\left(- 0.05 x^{3} + \left(0.5 x\right)^{0.5 x}\right) \left(0.1 x + \left(\frac{0.17}{{0.1}^{x} + 0.3}\right)^{x} - \frac{0.05}{x}\right)}{0.15 x + 0.98} - 0.06$ & $[0.08, 4.2]$ \\
8  & 43 & \makecell[l]{
    $\left(-0.02 + \frac{0.0061}{x}\right) \left|{x - 1}\right|^{4.54} + 0.348$ + \\
    $x \Big(\left(0.04 x + \frac{0.4}{x}\right) \Big(\left|{x^{0.66} - 1.2}\right|^{3.9} + \left|{x - 6.7}\right|^{-2.1}$ \\ $ - \frac{0.14}{x}\Big) - 0.08\Big)$
 } & $[0.5, 4]$ \\ 
9  & 39 & \makecell[l]{
  $0.339$ \\\scriptsize
  $ \left({0.73}^{{1.376}^{{1.095}^{x} {1.376}^{0.36 x^{1.661 - x}} \left(x - 1.7\right) \left(x^{x - 1.7} - 1.7\right)} \left(x^{x - 1.8} - 1.7\right)}\right)$
  }
  & $[0.5, 3]$
 \\
10 & 39 & $\left(0.943 x \left(x x^{x^{6.283}}\right)^{- x}\right)^{x^{0.723} x^{- 0.72 x^{9.87} \left(x \left(0.932 x\right)^{x}\right)^{- 9.9 x}}}$ & $[0.5, 1.5]$ \\
11 & 39 & \makecell[l]{
  $0.636 \left(x + 0.318\right)^{\left(0.05 x^{0.47 x - 0.745}\right)^{- x \left(1.4 \left({0.05}^{x^{2}}\right) - x\right)}} -$ \\
  $0.64 \left({0.05}^{0.745 - 0.47 x}\right)$ 
  }
  & $[0.5, 3.2]$ \\
12 & 37 & \makecell[l]{
  $\left(x - 1.6\right) \biggl(x^{{0.049}^{x - 1.6}} + \left(0.27 x^{0.073} - 0.028\right) $ \\ 
  $ \left(x^{{0.036}^{{0.104}^{x - 1.6} \left(2 x - 1.6\right)}} - 1.6\right) \biggr)$ 
  } 
  & $[0.2, 2]$ 
\end{tabular}
\end{table}

To verify the findings on synthetic data, we also run experiments on the one-dimensional Nikuradse dataset \cite{nikuradse1933stromungsgesetze}, which contains measurements of friction of fluids in pipes \cite{guimera2020bayesian}. Given the dense set of observations and its difficulty for modeling, this dataset has been used for testing SR implementations \cite{guimera2020bayesian,kronberger2026guiding} and allows us to verify whether the results hold in a more realistic setting. It contains 362 observations, which we randomly split into a training set of 230 and a test set of 132 observations \cite{kronberger2026guiding}. Although the noise level of the Nikuradse dataset is not explicitly specified, it is expected to be much higher than the noise level of the synthetic problems. In our experiments, we assume a noise level of $\sigma = 0.0367$ determined as the root mean squared error (RMSE) of well-fitting models found with some exploratory GP runs. We use the same settings shown in Table \ref{tab:operon_settings}, with a maximum complexity of 50.

\section{Results}
\label{sec:results}

\begin{figure}[t!]
\centering
\input{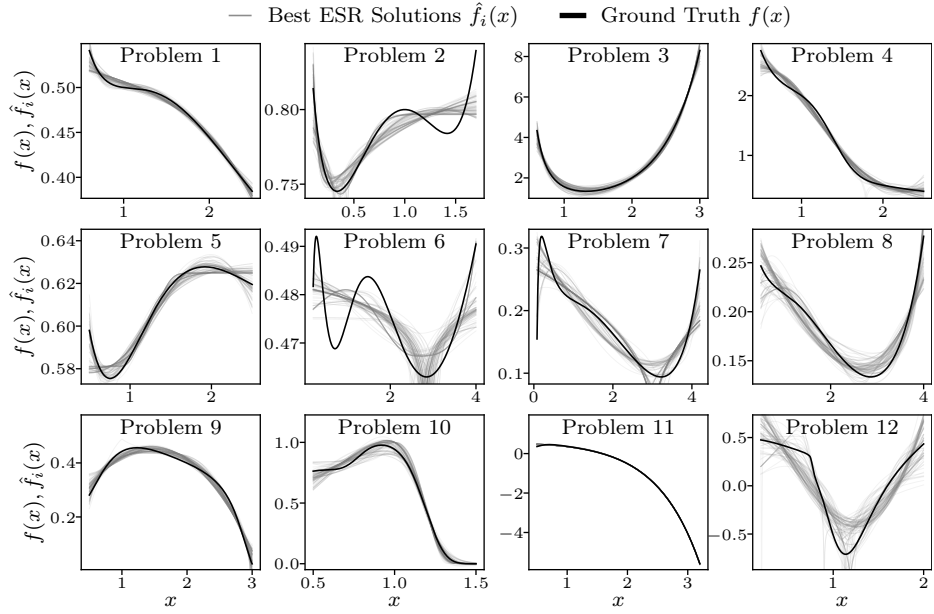}
\caption{Ground truth and 100 most accurate ESR solutions. In most problems, ESR finds models that capture the overall curvature of the ground truth but still have a high RMSE of between $10^{-2}$ and $10^{-1}$ compared to the noise level of $\sigma = 10^{-7}$.}
\label{fig:ground_truth_with_esr}
\end{figure}

Figure \ref{fig:ground_truth_with_esr} compares the best models found by ESR to the ground truth for each synthetic problem. Most ESR models capture the overall curvature, especially in Problems 1, 3 and 11. However, Problems 6, 7, and 12 show poor approximations. We report the root mean squared error (RMSE) in all our results which enables comparisons to the known irreducible noise level.
The RMSE of the best ESR models ranges between $10^{-2}$ and $10^{-1}$ in all synthetic problems, which is much higher than the noise level of $\sigma = 10^{-7}$, despite visual similarity of the outputs.

\begin{figure}[b!]
\centering
\input{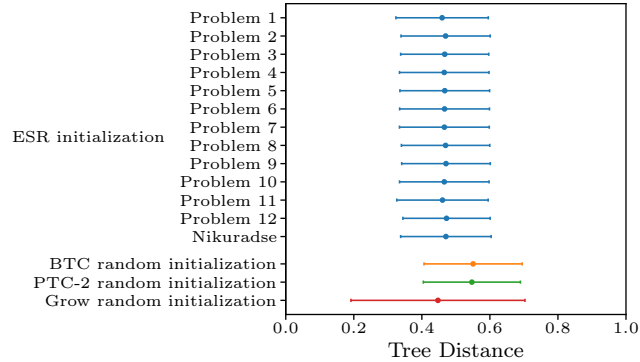}
\caption{The mean and standard deviation of pairwise tree distances among individuals in the initial population for each method. Higher values denote more diversity. As ESR initialization depends on the problem, the mean and standard deviation are shown for each problem separately. For the three random initialization methods, the bars show the mean and the average standard deviation of the initial populations of all 1000 runs. All methods provide similar diversity, with just slightly different averages and the grow method with higher standard deviations.}
\label{fig:diversity}
\end{figure}

As population diversity is crucial for the success of GP and ESR does not explicitly enforce dissimilarity within its search space, we first analyze whether a population initialized with ESR  provides sufficient diversity. To ensure that the subsequent evolutionary optimization is not compromised by a lack of diversity, we compare the diversity of an ESR-initialized population with randomly initialized populations using BTC, PTC-2 and the grow method. We use the hash-based tree distance \cite{burlacu2019hash} as a measure of syntactic dissimilarity between two expression trees. The hash-based tree distance is the ratio of distinct subtrees to the total number of subtrees in both expression trees, and also accounts for commutativity of multiplication and addition. Therefore, two trees that are equal or isomorphic regarding commutativity have a tree distance of zero, while two trees that do not share a single node have a tree distance of one.
Figure \ref{fig:diversity} shows the mean and standard deviation of pairwise tree distances between the models in the populations of each run with error bars for each method. Since the ESR-initialized population depends on the problem, we show the distribution for each problem separately. For each random population initialization method, we show the mean across all 1000 runs as well as the average standard deviation of each of the 1000 populations. Since the random initial population is independent of the problem, we show it only once.

The ESR-initialized populations show similar diversity across all problems. Their diversity is slightly below the diversity of populations initialized with BTC and PTC-2, but clearly within the standard deviation. The grow method shows slightly less diversity than the other methods with higher standard deviations, which is expected as it creates trees of any size and shape. Overall, the diversity of the ESR-initialized population is comparable to (although slightly lower than) the randomly initialized populations, suggesting that although the ESR-initialized population is more accurate, it is marginally less diverse which could hinder the evolutionary process.

\begin{figure}
\centering
\input{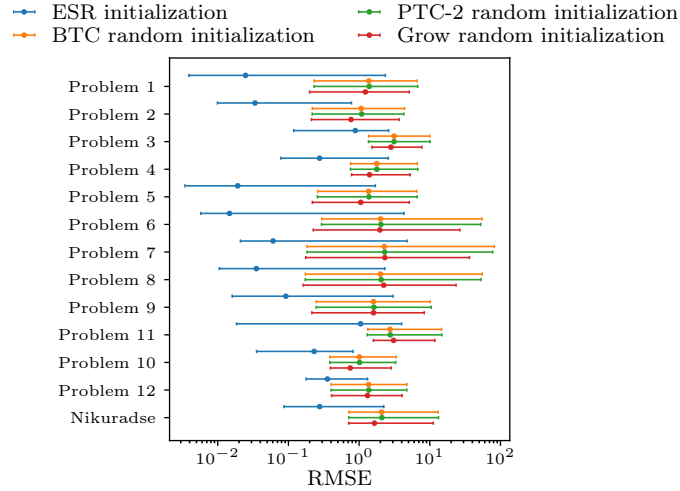}
\caption{The distribution of test errors of models in the initial population for all methods and problems. The dots show the median and the error bars denote the 5\textsuperscript{th} and 95\textsuperscript{th} percentile. The bars for the three random initialization methods show the average median and percentile values across all 1000 runs. As expected, we observe a much lower median test error in the ESR-initialized population than in the random initialization methods.}
\label{fig:initial_population_test_err}
\end{figure}

To quantify the accuracy advantage of the ESR-initialized population over the random initialization methods, we compare the distributions of test errors of models in the initial populations before any evolutionary optimization. Figure~\ref{fig:initial_population_test_err} shows the median and the centered 90\% of test errors for each initialization method and problem. Invalid outputs that can occur in randomly generated models are treated as infinite. 
For the random initialization methods, we report the average of the median and the 5\textsuperscript{th} and 95\textsuperscript{th} percentiles over all 1000 runs.
As expected, the ESR-initialized population achieves a median test RMSE one to two orders of magnitude lower than the random initialization methods, while still being clearly higher than the corresponding noise levels. The three random initialization methods yield very similar error distributions.

\begin{figure}[t!]
\centering
\input{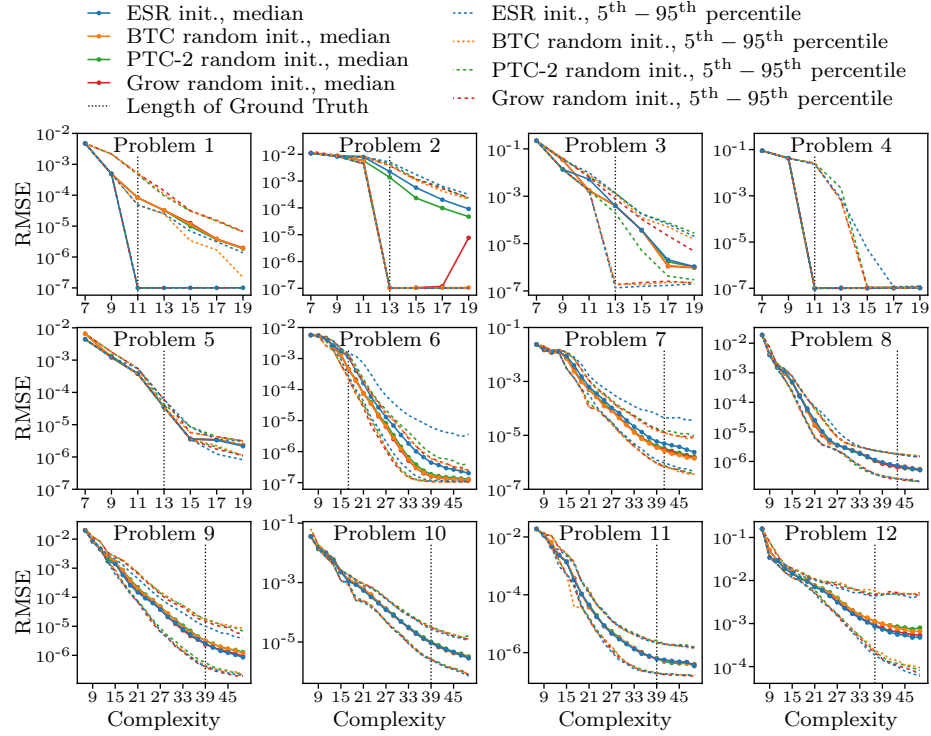}
\caption{
The distribution of test error over complexity of the Pareto-optimal models from each of the 1000 runs per initialization method for the synthetic problems. The areas between the dashed lines denote centered 90\% of test errors and the solid line the median at a specific complexity. All methods show similar performance with very similar distributions denoted by largely overlapping areas between the dashed lines. ESR initialization shows slightly better performance only in Problem~1 but performs even slightly worse in Problems~6 and 7. 
}
\label{fig:pareto_front_test_err}
\end{figure}

To compare the error distributions of the final GP results of all four methods, we take the models from the Pareto front after 500 generations of all 1000 GP runs for each method and problem. Figure \ref{fig:pareto_front_test_err} shows the distribution of test error over complexity for each problem. The area between the dashed lines 
denotes the centered 90\% of the test error of models at a specific complexity across all 1000 runs for a single method. The solid lines denote the median test error, and each dot indicates a specific complexity value. Given that ESR results are optimal with respect to accuracy within its search space, the dot at complexity 9 for GP with an ESR-initialized 
population shows the accuracy of the best ESR model. In all problems, this RMSE is between $10^{-2}$ and $10^{-1}$. We omit the training error results as they are nearly identical to the shown test results.

Despite the observed differences in the initial population with respect to diversity and accuracy, 
all methods perform nearly identically after the evolutionary process, with largely overlapping error distributions. Even the grow method with its slightly lower diversity and higher standard deviation performs equally well as the other methods. The ESR-initialized population also shows no clear improvement over GP with random population initialization and even performs worse in Problems 6 and 7. A visible improvement occurs only in Problem 1, where we reliably identify the ground truth with an ESR-initialized population.

\begin{figure}[t!]
\centering
\input{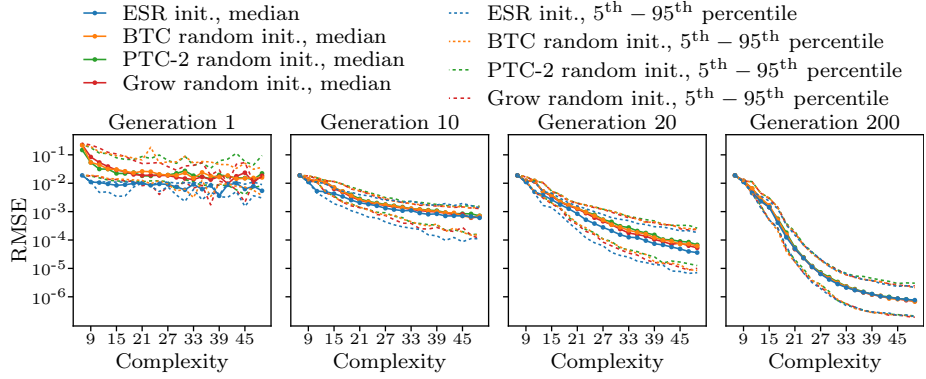}
\caption{The distribution of test error over complexity of the Pareto-optimal models from each of the 1000 runs for each initialization method in the synthetic Problem 11 after four different numbers of generations. While the ESR-initialized population performs better initially, these differences shrink continuously after ten and 20 generations until they perform equally after 200 generations.}
\label{fig:pareto_fronts_over_generations_eq_11}
\end{figure}

\begin{figure}
\centering
\input{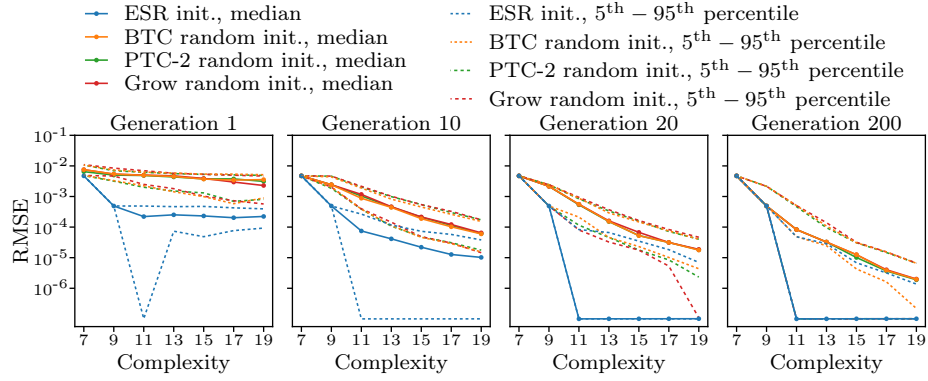}
\caption{The distribution of test error over complexity of the Pareto-optimal models from each of the 1000 runs for each initialization method in the synthetic Problem 1 after four different numbers of generations. In contrast to all other problems, the ESR-initialized population shows a clear improvement over the randomly initialized population, which is due to the best ESR model being very close to the ground truth.}
\label{fig:pareto_fronts_over_generations_eq_1}
\end{figure}

In most problems, such as Problem 11,
all error distributions overlap completely. While the ESR-initialized population provides slightly better results after approximately the first twenty generations, the randomly initialized populations catch up quickly. Figures \ref{fig:pareto_fronts_over_generations_eq_11} and \ref{fig:pareto_fronts_over_generations_eq_1} show the test error distributions of models from the Pareto fronts of all runs after different generation counts for each initialization method for Problems 11 and 1, respectively. Figure \ref{fig:pareto_fronts_over_generations_eq_11} shows for Problem 11 that after generation one, GP with ESR initialization shows better results, but after ten generations, both error distributions largely overlap, with the ESR-initialized population showing only slightly better results. This difference shrinks further after 20 generations. After 200 generations, both methods show identical performance. This pattern is observed in all problems except Problem 1, which we discuss in more detail below. Many models created by the grow method exceed the complexity limits. Such models are not explicitly handled in Operon and appear on the Pareto front in the first few generations until dominated during the evolutionary process. We removed these models in early generations from our analysis as they would distort the results.
We note that ESR initialization, which takes roughly two hours for all complexities per problem on an AMD EPYC 7713P 2.0 GHz 64-core CPU, is computationally much more expensive 
than the subsequent GP run, which took around only five seconds on the same CPU. These differences in computational effort 
outweigh any slight improvement in runtime of the GP run.

The only clear improvement in accuracy and complexity by the ESR initialization is observed in Problem 1. We attribute this to the ESR solution $\hat{f}(x) = \frac{1}{x + x^{0.0974 - x}}$ being very close to the ground truth $f(x) = \frac{1}{x + x^{{0.05}^{x} - x}}$, and therefore only requiring minor modifications by the evolutionary process. This is consistent with the observed search progress by GP shown in Figure \ref{fig:pareto_fronts_over_generations_eq_1}, where GP with an ESR-initialized population identifies the ground truth after only around 20 generations in most runs. In contrast, GP with random initialization methods identifies the ground truth much less frequently even after 500 generations as shown in Figure \ref{fig:pareto_front_test_err}. 

To ensure results are not affected by the specific algebraic representation of models selected by ESR, which uses symbolic simplification to prevent duplicates \cite{bartlett2023exhaustive}, we also run experiments with randomly sampled algebraic representations of the selected ESR models. These experiments show nearly identical results, with no significant improvement of the ESR-initialized population over random initialization in most problems.

\begin{figure}
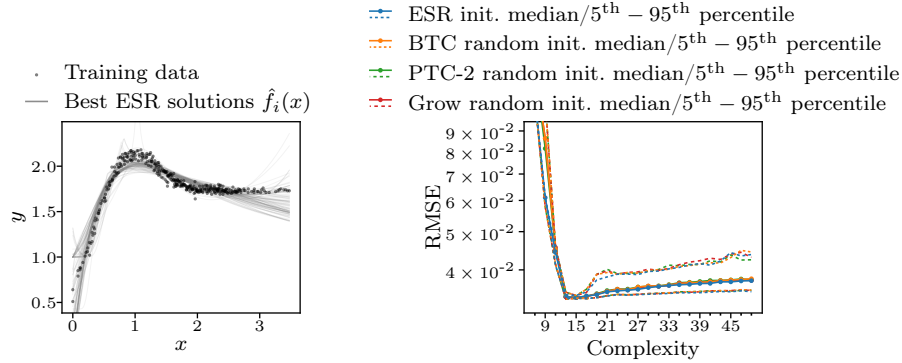

    \centering
    \subfloat[The Nikuradse data and the 100 most accurate ESR models. The best ESR model achieves an RMSE of 0.077, around twice as high as the assumed noise level of $\sigma = 0.03672$.\label{fig:nikuradse_data}]{%
        \input{plots/nikuradse_data.pgf}\hspace{1em}
    }\hspace{1em}%
    \subfloat[The distribution of test error over complexity of the Pareto-optimal models from each of the 1000 runs per initialization method \cite{nikuradse1933stromungsgesetze,guimera2020bayesian}. The areas between the dashed lines denote centered 90\% of test errors and the solid line denotes the median at a specific complexity across all 1000 runs. \label{fig:pareto_front_test_err_nikuradse}]{%
        \input{plots/final_pareto_fronts_nikuradse_1d.pgf}
    }
    \caption{The one-dimensional real-world Nikuradse dataset \cite{nikuradse1933stromungsgesetze,guimera2020bayesian} and the GP results for each initialization method, in which we observe a similar pattern as in the synthetic problems. While ESR-initialization provides reasonable fits in the initial population, as shown in Figure \ref{fig:nikuradse_data}, the GP yields similar results over all initialization methods, as denoted by the overlapping error distributions in Figure \ref{fig:pareto_front_test_err_nikuradse}.}
    \label{fig:nikuradse_data_results}
\end{figure}

The results on the real-world Nikuradse dataset in Figure \ref{fig:nikuradse_data_results} show the same pattern as in the synthetic problems and verify that this behavior is also consistent in a setting containing more noise. Figure \ref{fig:nikuradse_data} plots the whole dataset as well as the curves of the 100 most accurate ESR models. It shows that the best ESR models provide only rough estimates with RMSE values clearly above the assumed noise level of $\sigma = 0.03672$. Figure \ref{fig:pareto_front_test_err_nikuradse} shows that all methods provide similar results after running GP for 500 generations with largely overlapping error distributions of the final Pareto fronts.

\begin{figure}[t!]
\centering
\input{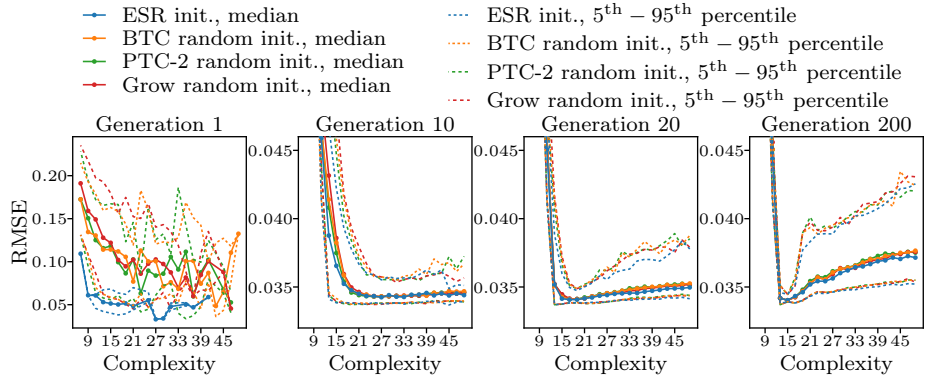}
\caption{The distribution of test error over complexity of the Pareto-optimal models from each of the 1000 runs for both methods in the Nikuradse dataset \cite{nikuradse1933stromungsgesetze,guimera2020bayesian} after four different numbers of generations. As in the synthetic problems, the advantage of the ESR-initialized population diminishes after only a few generations and does not lead to more accurate or less complex models.}
\label{fig:pareto_fronts_over_generations_nikuradse}
\end{figure}

Figure \ref{fig:pareto_fronts_over_generations_nikuradse} also shows that, for this dataset, the initial advantage of an ESR-initialized population quickly diminishes after a few generations and does not lead to more accurate or less complex models. The same observation applies to all three random initialization methods, which perform equally well.

\section{Conclusion}
\label{sec:conclusion}

Despite the intuitive assumption that optimizing the initial population for accuracy and complexity would improve the Pareto front of models in GP/SR, we find that an ESR-initialized population does not provide significant improvements over random initialization. Even with the best ESR population within the tested search space 
as GP's initial population, GP with a randomly initialized population is able to recover any information provided by ESR on its own in the final Pareto front and quickly catches up with GP with an ESR-initialized population. Only in one problem, where the ground truth is very close to the best ESR model, do we observe an improvement. In all other problems, the initial advantage quickly diminishes within only a few generations. The same observation applies to the three random initialization methods, which all perform equally well. The grow method and its limited control over the complexity of the initial population led to models that exceed the given complexity limits and provided slightly less and more varying diversity. However, even these aspects do not impair the final results. This implies that the specific choice of initialization method barely has an effect on the final results in GP/SR. 
We recommend BTC or PTC-2 because they provide freely customizable complexity limits, are computationally inexpensive and lead to equivalent results in the subsequent evolutionary process as the grow method or ESR initialization.

A limitation of our study is the small complexity in the optimized, ESR-initialized population. Although ESR provides optimal results within its supported complexity range, the error and complexity of the best ESR models are still far off from the noise level and the complexity of the ground truth. Since we observe a clear improvement in Problem 1 where the ground truth and the best ESR model are syntactically close, larger models in the initial population that are optimal in their covered search space might lead to more improvements over random initialization and increase the chance of finding a better or even optimal solution with GP.
Even if such models in the larger-complexity initial population were not optimal, perhaps some `reasonable' guesses based on domain-specific knowledge could be beneficial, as one would expect that minor modifications to these could achieve good solutions.

We have only studied univariate problems. With higher-dimensional problems there may be differences between the random initialization methods. However, ESR cannot be used as it is limited to low dimensionality. 
Additionally, our results are based mostly on synthetic problems and only one real-world dataset. Other real-world problems might offer different possibilities for improvements, such as biasing the population towards specific functional forms based on prior knowledge about the underlying system.
Finally, the complexity of the test functions also plays a role. We selected test functions with a range of complexities but potentially the datasets can be approximated well with short expressions which could make it easy for GP to evolve fitting expressions, even when the generating function is not recovered exactly.

This work tests the effect of adapting only the initial population using ESR on the final Pareto front after many generations. Since we do observe beneficial effects of initialization, other approaches that maintain a stronger influence throughout the whole evolutionary process, such as periodic injection of elite individuals from ESR, might still lead to improvements over random initialization.

\subsubsection{\ackname}
{\small
L.K. and G.K. acknowledge support by the Austrian Federal Ministry for Economy, Energy, and Tourism, the Federal Ministry for Innovation, Mobility and Infrastructure, and the regional government of Upper Austria within the COMET project ProMetHeus (904919) supported by the Austrian Research Promotion Agency (FFG). D.J.B. acknowledges that support was provided by Schmidt Sciences, LLC. H.D. is supported by a Royal Society University Research Fellowship (grant no. 211046).
}

\bibliographystyle{splncs04}
\bibliography{references}

\end{document}